\pdfoutput=1
\documentclass[11pt]{article}

\usepackage[final]{acl}

\usepackage{times}
\usepackage{latexsym}
\usepackage{booktabs}

\usepackage[T1]{fontenc}
\usepackage[utf8]{inputenc}

\usepackage{microtype}

\usepackage{inconsolata}

\usepackage[textsize=tiny]{todonotes}
\usepackage{makecell, siunitx}
\usepackage{array}
\usepackage{multicol}
\usepackage{multirow}
\usepackage{enumitem}
\usepackage{xurl}

\title{LLMTaxo: Leveraging Large Language Models for Constructing Taxonomy of Factual Claims from Social Media}

\author{Haiqi Zhang, Zhengyuan Zhu, Zeyu Zhang, Chengkai Li \\
        University of Texas at Arlington \\
        \texttt{\{haiqi.zhang, zhengyuan.zhu, zeyu.zhang\}@mavs.uta.edu}\\ \texttt{cli@uta.edu}}

\begin{document}
\newcommand{\system}[1]{{\small \ensuremath {\mathsf{#1}}}}
\newcommand{\LLMTaxo}{\system{LLMTaxo}}
\maketitle
\begin{abstract}
With the rapid expansion of content on social media platforms, analyzing and comprehending online discourse has become increasingly complex. This paper introduces \LLMTaxo, a novel framework leveraging large language models for the automated construction of taxonomies of factual claims from social media by generating topics at multiple levels of granularity. The resulting hierarchical structure significantly reduces redundancy and improves information accessibility. 
We also propose dedicated taxonomy evaluation metrics to enable comprehensive assessment. 
Evaluations conducted on three diverse datasets demonstrate \LLMTaxo's effectiveness in producing clear, coherent, and comprehensive taxonomies.  
Among the evaluated models, GPT-4o mini consistently outperforms others across most metrics. The framework’s flexibility and low reliance on manual intervention underscore its potential for broad applicability. 
\end{abstract}

\section{Introduction}
\label{sec:introduction}

Misinformation, also known as false or misleading information~\cite{wu2019misinformation}, has the potential to sway public perception, cause confusion, and influence people's decision-making processes~\cite{del2016spreading, muhammed2022disaster}. Social media platforms, in particular, facilitate the rapid sharing of vast amounts of content, blending accurate information with falsehoods~\cite{allcott2019trends}. Social media’s global reach and ease of use have transformed how millions of users exchange opinions, news, and factual claims in real-time, making it fertile ground for misinformation~\cite{aimeur2023fake}. 
Factual claims, which are assertions that can be verified as either true or false, are a common vehicle for misinformation~\cite{ni2024afacta}. 
These claims, whether accurate or not, have a profound societal impact, as the public tends to believe a factual claim is true regardless of its truthfulness~\cite{moravec2018fake, ognyanova2020misinformation, xiao2021dangers, ratsd-compjour22, zhang-climatenlp24, ratsd-naacl25}.

The dynamic nature of social media often leads to repetitive or reformulated claims, complicating the identification and validation of factual content~\cite{zhou2015impact}, making it difficult for non-technical individuals and researchers from different fields to navigate the vast amounts of data~\cite{suarez2021prevalence, hook2022social, muhammed2022disaster}.
This challenge calls for scalable and automated tools to systematically organize and analyze factual claims, helping stakeholders---such as researchers and fact-checkers---more effectively navigate the complex information landscape.

One promising approach to addressing this challenge is the use of taxonomies, which provide structured systems for organizing complex information. Taxonomy has been applied in misinformation categorization. 
For example, \citet{tambini2017fake} classified fake news into distinct categories to capture the diverse forms and manifestations of the phenomenon. 
\citet{kumar2018false} constructed taxonomies of false information on social media based on various characteristics, such as whether the content is intended to deceive and whether it is opinion-based or fact-based.  \citet{zhao2022people} proposed a taxonomy of misinformation on social media based on falsity level and evidence type. 
Yet, these approaches do not construct topic taxonomies, which are particularly valuable for exploring and organizing factual claims.
In practice, many fact-checking websites, such as PolitiFact,\footnote{\url{https://www.politifact.com/}} Snopes,\footnote{\url{https://www.snopes.com/}} and Full Fact,\footnote{\url{https://fullfact.org/}} use topics to organize fact checks. For example, the claim ``\textit{Right to Try experimental drug program saved thousands and thousands of lives}'' in PolitiFact is given topics such as ``Elections'' and ``Health Care.'' However, these topics are too broad to help users efficiently locate specific claims. 

Motivated by the challenges, prior studies, and practical applications discussed above, this paper introduces \LLMTaxo, a novel framework utilizing large language models (LLMs) for automatically constructing a topic taxonomy given a collection of factual claims from social media within a specific topic domain. 
The taxonomy starts with broad categories and branches into increasingly specific subcategories. 
In this way, factual claims can be categorized into hierarchical categories, allowing for the efficient exploration of information at multiple levels of granularity. For example, broad categories can group COVID-19 vaccine-related claims into overarching topics such as ``Public Health,'' while more detailed categories can address specific claims about ``Vaccine Safety and Effectiveness.'' 

Our framework, \LLMTaxo, organizes factual claims into a coherent hierarchical taxonomy by clustering semantically similar claims, identifying distinct ones, and generating topics at multiple levels of granularity---broad, medium, and detailed.
Leveraging the rich background knowledge of LLMs and few-shot learning, \LLMTaxo\ 
minimizes human involvement and automates the categorization process. It effectively addresses challenges such as semantic variability (the expression of similar ideas in diverse ways) and ensures scalability and adaptability to social media discourse across various datasets and topic domains.

To demonstrate the generalizability of \LLMTaxo, we conducted evaluations using carefully designed metrics across three distinct datasets spanning different topic domains and data sources. These include social media posts from \textit{X} (formerly Twitter) and Facebook, covering domains such as COVID-19 vaccines, climate change, and cybersecurity.
The evaluation results, showing all models scoring above 3.0 out of 5 across all metrics,  with a highest score of 4.9, demonstrate the effectiveness of \LLMTaxo\ in generating clear and comprehensive taxonomies. 
By identifying distinct factual claims, the framework substantially reduces redundancy. 
The hierarchical structure of the taxonomy produced by \LLMTaxo\ allows users to explore claims at multiple levels of detail. 
Moreover, \LLMTaxo\ exhibits consistently strong performance across diverse datasets,  demonstrating its adaptability and potential for wide-ranging applications.

In summary, this paper makes several key contributions:
\begin{itemize}[noitemsep,wide,topsep=0pt]
    \item We introduce the first multi-level taxonomy of factual claims from social media constructed using LLMs. This taxonomy can be integrated into fact-checking workflows and applied across various research domains. 
    \item We are the first to utilize LLMs for generating multi-granularity topics.
    \item We develop a set of evaluation metrics for comprehensively assessing the structural quality of taxonomies and their claim-topic alignment. 
    \item We evaluate our taxonomies across three diverse datasets. The results demonstrate \LLMTaxo's adaptability to different domains while maintaining accuracy in constructing meaningful taxonomies.
    \item We release a curated dataset and a public codebase at \url{https://github.com/idirlab/LLMTaxo}. They are valuable for supporting both future research and reproducibility.    
\end{itemize}

\section{Related Work}

\paragraph{Taxonomy Construction.}
Although taxonomy construction has been extensively studied, the definitions of specific problems vary. Generally, taxonomies are hierarchically structured classifications of concepts, terms, and entities that help users organize, retrieve, and navigate information~\cite{carrion2019taxonomy,yang2012constructing}. 
Generic taxonomy construction tasks typically involve short concept terms or entity names, which are often structured as hypernym-hyponym pairs~\cite{zhang2018taxogen, huang2020corel}. 

Constructing taxonomies from broader, less formatted content, such as social media posts, differs from traditional taxonomy construction. The inherent variability of such content makes it challenging to establish a precise taxonomy.
Several studies have attempted to address this problem. \citet{durham2023unveiling} explored automatic taxonomy generation from disaster-related tweets using topic modeling techniques~\cite{blei2003latent, dumais2004latent}. \citet{najem2021semi} proposed semi-automatic ontology construction of tweets based on semantic feature extraction using WordNet and BabelNet~\cite{WordNet, navigli2010babelnet}.

The rise of LLMs has significantly advanced many natural language processing tasks, but limited effort has been devoted to taxonomy construction.  \citet{chen2020constructing} employed pretrained language models to construct taxonomic trees, while \citet{chen2023prompting} compared prompting and fine-tuning approaches for hypernym taxonomy construction. Additionally, \citet{shah2023using} and \citet{wan2024tnt} introduced end-to-end pipelines that integrate LLMs to generate, refine, and apply labels for user intent analysis in log data. 
Our work extends this line of research by applying LLMs to generate multi-level taxonomies for factual claims and social media posts. Like prior efforts, we harness LLMs' generative and semantic capabilities, but we focus on taxonomy construction in open-domain, noisy, and high-variance contexts where entity boundaries and hierarchical relationships are less clear.

\paragraph{Topic Generation.}
LLMs have emerged as a promising alternative to traditional topic generation approaches. They can generate topics from a given set of documents without requiring predefined labels or training data~\cite{sarkar2023zero}. This capability allows for more flexible and context-aware topic extraction. For instance, \citet{mu2024large} introduced a framework that prompts LLMs to generate topics and adhere to human guidelines for refining and merging topics. 
Recent research has also explored different LLM-based topic modeling techniques. For example, BERTopic~\cite{grootendorst2022bertopic} has shown superior performance in terms of diversity and coherence across multiple datasets~\cite{jung2024expansive}. Yet, to the best of our knowledge, no existing work has generated topics at multiple levels of granularity.

\section{Methodology}

Analyzing social media data presents numerous challenges, including the overwhelming volume of content with high repetition and the substantial human effort required to explore the data. 
Our \LLMTaxo\ framework is designed to systematize online content through automated construction of a taxonomy of factual claims.  This taxonomy serves as a hierarchical classification system. It enhances the accessibility and navigability of information for users by categorizing claims into broad, medium, and detailed topics. 

\begin{figure*}[ht]
    \begin{center}    	\includegraphics[width=2.1\columnwidth]{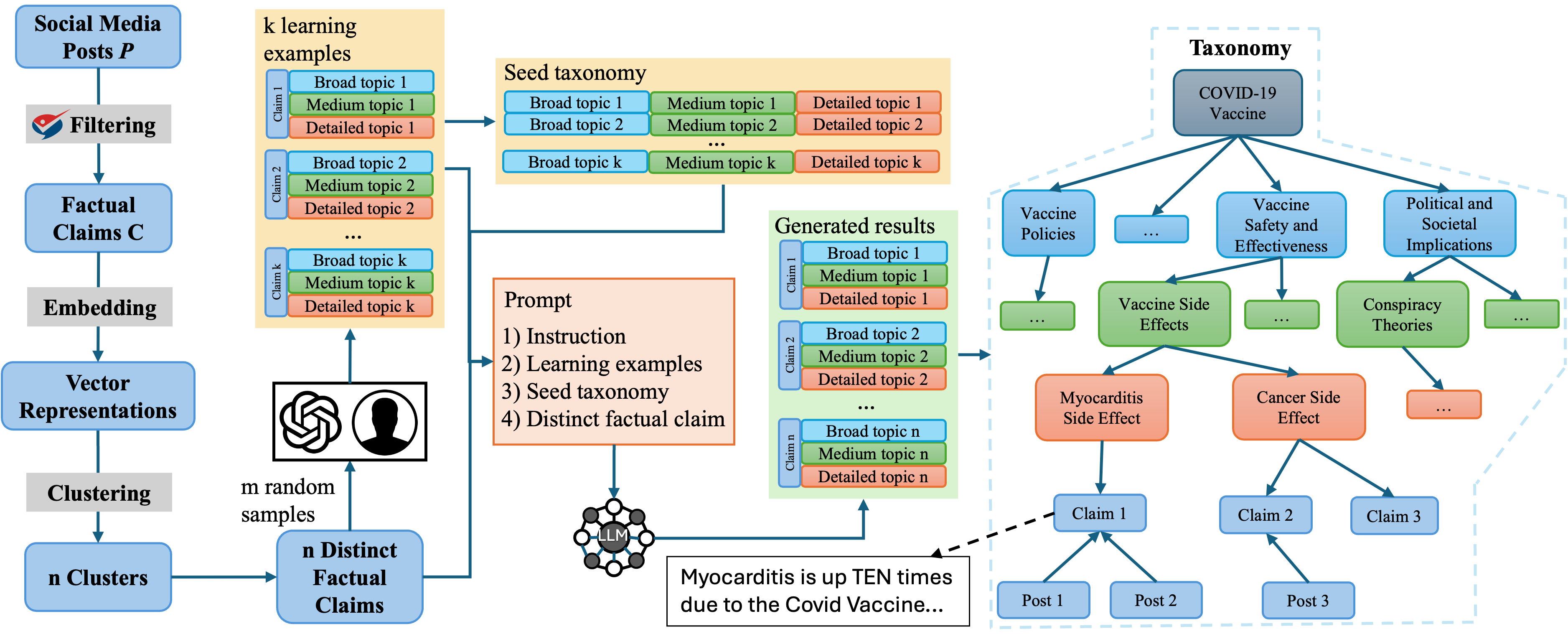}
    \caption{The \LLMTaxo\ framework.}
    \label{fig:framework}
    \end{center}
    \vspace{-2mm}
\end{figure*}

\LLMTaxo\ initially identifies factual claims from social media posts, and subsequently clusters similar claims to discern and select distinct ones, thereby reducing redundancy. It then leverages LLMs to generate topics for each distinct claim at multiple levels of granularity, ultimately constructing a structured taxonomy.
The overview of our framework is shown in Figure~\ref{fig:framework}.

\subsection{Taxonomy Construction}
\label{sec:taxonomy_construction}
Our primary goal is to construct a hierarchical taxonomy. The hierarchical design is informed by the prototype theory~\cite{geeraerts2006prototype}, which guides cognitive categorization, i.e., the three-level hierarchy of the taxonomy. Formally, a hierarchical taxonomy $\mathcal{T}=(T_{b}, T_{m}, T_{d}, f_{m}, f_{d})$ contains topics at multiple levels of granularity, ranging from broader to more fine-grained topics. Specifically, the taxonomy uses a three-tiered structure, consisting of broad topics $T_{b}$ (representing general themes), medium topics $T_{m}$ (reflecting intermediate distinctions), and detailed topic $T_{d}$ (highlighting finer aspects). 
The functions $f_m$ and $f_d$ define the hierarchical relationships between topics, where $f_{m}: T_{m}$$\rightarrow$$T_{b}$ maps each medium topic to its corresponding broad topic, and $f_{d}: T_{d}$$\rightarrow$$T_{m}$ maps each detailed topic to its respective medium topic. 

Note that the same topic label may appear under different parent topics, but these instances represent conceptually distinct topics. For example, the medium topic ``Vaccine Safety'' under broad topic ``Public Opinion'' is essentially ``Public Opinion about Vaccine Safety,'' which is meaningfully different from ``Vaccine Safety'' under ``Government Policies.'' To minimize redundancy while preserving clarity, we use concise topic labels for medium and detailed topics, as their inherent meanings are informed by their parent topics. 

Given a collection of factual claims $\mathcal{C} = \{c_1, c_2, ..., c_n\}$,  each claim $c \in C$ is assigned a tuple of topics at three levels---broad, medium, and detailed---with a mapping function $\phi$. More formally, $\phi(c)=(t_{b}, t_{m}, t_{d})$ where $t_{b}\in T_{b}$, $t_{m}\in T_{m}$, and $t_{d}\in T_{d}$. Note that the detailed topic is optional because some claims may be too brief or ambiguous to be categorized distinctly at all three levels. 

For example, the right side of Figure~\ref{fig:framework} illustrates the topic tuple for a claim related to COVID-19 vaccine: ``\textit{Myocarditis is up TEN times due to the Covid Vaccine...}'' The broad topic $t_{b}$ is ``Vaccine Safety and Effectiveness,'' the medium topic $t_{m}$ is ``Vaccine Side Effects,'' and the detailed topic $t_{d}$ is ``Myocarditis Side Effect.''

To automate the process of taxonomy construction given a collection of factual claims $\mathcal{C}$, we create learning examples that contain both sample claims and their corresponding topics. The topics from the learning examples form a seed taxonomy, which serves as a foundation for LLM-based expansion. To generate topics for each distinct factual claim in $\mathcal{C}$, 
we prompt the LLMs with both the learning examples and the seed taxonomy. After generating topics for all claims, we consolidate them to construct a refined taxonomy.  This process is illustrated in Figure~\ref{fig:framework}. 

\subsubsection{Learning Examples and Seed Taxonomy}
\label{sec:learning_examples}
While LLMs possess broad general knowledge, they are not pretrained to directly generate hierarchical topics for factual claims at different granularities. Our initial experiments showed that LLMs often generate inconsistent topics for similar claims, resulting in an excessive and unwieldy number of topics that hinder comprehension, as further detailed in Section~\ref{sec:ablation}. 
To mitigate this issue, we propose to expand and refine a seed taxonomy $\mathcal{S}$ derived from a set of sample factual claims. This initial taxonomy aids the LLMs by providing a foundation to expand upon. We leverage an LLM with human-in-the-loop supervision to create $k$ learning examples $\mathcal{L}$$=$$\{\langle c_{1}, (t_{b}^{(1)}, t_{m}^{(1)}, t_{d}^{(1)})\rangle, ..., \langle c_{k}, (t_{b}^{(k)}, t_{m}^{(k)}, t_{d}^{(k)})\rangle\}$, representing claims with their respective topics at each level of granularity. The 3-level topic tuples from the $k$ claims form the seed taxonomy, denoted as $\mathcal{S}$$=$$\{(t_{b}^{(1)}, t_{m}^{(1)}, t_{d}^{(1)}), ..., (t_{b}^{(k)}, t_{m}^{(k)}, t_{d}^{(k)})\}$. These learning examples and the seed taxonomy facilitate few-shot in-context learning, helping to stabilize the variation and number of topics generated. The size $k$ of learning examples varies across datasets.

To compile the learning examples and the seed taxonomy, we randomly select a subset $\mathcal{R}$$\subset$$\mathcal{C}$ containing $m$ distinct claims for annotation. Recognizing that direct human annotation is labor-intensive and the annotation requires high accuracy and consistency, we employ GPT-3.5~\cite{brown2020language} to assist human in topic annotation for these claims.

Each claim $c \in \mathcal{R}$ is processed by an LLM prompted to generate topics at three levels of granularity. Below is an example of the prompt used for the COVID-19 vaccine topic domain.

\vspace{2mm}
{\small\fontfamily{cmtt}\selectfont 
You will be given a tweet related to COVID-19 vaccine. Please generate topics for the tweet from different granularities such as broad topic, medium topic, and detailed topic. Each generated topic should be no more than eight words and you should try to minimize the number of topics generated.

Examples: ...
}
\vspace{2mm}

\LLMTaxo\ adopts a multi-round human-in-the-loop annotation process. In each round, the LLM generates topics for a batch of 10 claims. 
These topics are then evaluated and refined by annotators, resulting in a finalized annotation of each claim $c$ with topic tuple $(t_{b}, t_{m}, t_{d})$, forming a claim-topic pair $\langle c, (t_{b}, t_{m}, t_{d})\rangle$. 
The LLM may produce diverse topic labels in the initial run. To ensure consistency and minimize topic variation, we restrict the number of distinct topics and standardize topic labels across similar claims during the refinement process. For example, if the LLM generates “COVID-19 Vaccine Mandates” for claim A and “Vaccine Mandates” for claim B, the annotators are expected to retain a single, generalized topic (e.g., “Vaccine Mandates”) for both claims. 

The refined claim-topic pairs are subsequently used as in-context examples to guide the LLM in the next round. This process is iterated until topics are generated for all $c \in \mathcal{R}$.
The human-in-the-loop process reduces manual effort and allows for tailored annotations based on specific requirements, which is adaptable across various topic domains.

Upon annotating all $m$ claims, the annotators review the most frequent broad topics and manually pick $k$ representative claim-topic pairs as the learning examples $\mathcal{L}$. 
The topics from these $k$ examples form the seed taxonomy $\mathcal{S}$.   
The seed taxonomy and the learning examples guide the LLMs to generalize the task of topic generation across a diverse range of claims while ensuring consistent and structured outputs.

\subsubsection{Multi-level Topic Generation}
To automate the taxonomy construction process and minimize the human effort, we employed LLMs to generate topics $(t_{b}, t_{m}, t_{d})$ for each factual claim $c \in \mathcal{C}$. 
The learning examples $\mathcal{L}$ and the seed taxonomy $\mathcal{S}$ are utilized as part of the prompt for the LLM. 
Specifically, the prompt consists of $l_{i} \in \mathcal{L}$ for $i = 1, 2, \dots, k$ (where $l_i = \langle c_i, (t_{b}^{(i)}, t_{m}^{(i)}, t_{d}^{(i)})\rangle$ is the claim-topic pair of the $i$-th learning example), the seed taxonomy $\mathcal{S}$, the instruction and questions that ask the LLM to produce the topic tuple for the claim $c$. 
Due to the limited context length of LLMs, each prompt generates the topic tuple $\phi(c)=(t_{b}, t_{m}, t_{d})$ for only one $c$. 
This generation process is iterated until finishing generating topic tuples for all $c \in \mathcal{C}$. 
The prompt is detailed in Figure~\ref{fig:prompt} in Appendix~\ref{sec:prompt}.

\subsubsection{Topic Consolidation}
After the LLM generates the topic tuples for all claims, \LLMTaxo\ consolidates the results to build the taxonomy. Given that $(t_{b}, t_{m}, t_{d})$ are generated as the topic tuple for a claim, $t_b$, $t_m$, and $t_d$ inherently form parent-child relationships. Medium topics that align with the same broad topic are treated as child nodes of that broad topic, and detailed topics are similarly considered child nodes of their respective medium topics. 
Formally, from the generated topic tuples, \LLMTaxo\ derives the two mapping functions $f_{m}$$: T_{m}$$\rightarrow$$T_{b}$ and $f_{d}$$: T_{d}$$\rightarrow$$T_{m}$. 
$f_m$ maps each medium topic to a broad topic: $f_m\left(t_m\right)=t_b$ if $\exists\ c$$\in$$\mathcal{C}$ such that $\phi(c)$$=$$(t_b, t_m, \rule{0.3cm}{0.2mm})$. 
Similarly, $f_d$ maps each detailed topic to a medium topic: $f_d\left(t_d\right)=t_m$ if $\exists\ c$$\in$$\mathcal{C}$ such that $\phi(c)$$=$$(\rule{0.3cm}{0.2mm}, t_m, t_d)$.  

For instance, consider two claims $c_1$ and $c_2$ with the generated topic tuples $\phi(c_1)$ = (\textit{Vaccine Safety and Effectiveness}, \textit{Vaccine Side Effects}, $t_d^{(1)}$) and $\phi(c_2)$ = (``Vaccine Safety and Effectiveness,'' ``Vaccine Injury,'' $t_d^{(2)}$). The derived mapping function $f_{m}$ will include $f_{m}$(``Vaccine Side Effects'') = ``Vaccine Safety and Effectiveness'' and $f_{m}$(``Vaccine Injury'') = ``Vaccine Safety and Effectiveness.'' 
This establishes that ``Vaccine Side Effects'' and ``Vaccine Injury'' are child nodes of ``Vaccine Safety and Effectiveness.''

\subsection{Claim Detection}
\label{sec:claim_detection}
Social media contains a wide range of content, including personal opinions, personal experiences, and entertainment. Prior to taxonomy construction, we apply claim detection to identify social media posts that are more likely to contain factual claims, which could potentially carry misinformation.
We employ the ClaimBuster~\cite{claimbuster-kdd17} model which assigns a score to each sentence or paragraph, indicating the likelihood of it being a check-worthy factual claim. A higher score suggests a greater likelihood of check-worthiness.
We set a threshold of 0.5 as it effectively balances precision and recall in identifying check-worthy claims. Posts that score above this threshold are retained for further steps of \LLMTaxo.
This approach reduces the dataset to a more manageable subset (see Section~\ref{sec:dataset} for details). It also ensures that the retained posts are more likely to contain check-worthy factual information relevant to our taxonomy construction.

\subsection{Identifying Distinct Claims}
Many factual claims on social media are frequently repeated or rephrased. For example, the posts ``\textit{BREAKING: Pentagon rescinds COVID-19 vaccine mandate}'' and ``\textit{The Pentagon officially rescinds COVID-19 vaccine mandate}'' convey the same claim but are phrased slightly differently. 
To reduce redundancy and focus on unique claims, we apply clustering to group identical or nearly identical factual claims. We use HDBSCAN~\cite{campello2013density} due to its ability to handle noise and detect outliers, which is particularly useful given that many posts do not closely resemble others.
To capture the semantic meaning of claims, we employ Sentence-BERT~\cite{reimers2019sentence} to generate dense vector representations. After clustering, we identify distinct claims by selecting the first post from each returned cluster list while excluding the outlier cluster.
A distinct claim is the representation of a cluster of similar factual claims, and the identified distinct claims~$\mathcal{C}$ are used for taxonomy construction. The outliers represent infrequently discussed content, whereas we only focus on content that appears multiple times. 
Repeated exposure to information increases belief in its accuracy~\cite{pennycook2018prior}, underscoring the significance of identifying widely circulated claims.

\section{Experiments}

\subsection{Datasets}
\label{sec:dataset}

To evaluate our method, we conducted experiments on three social media datasets, each covering a specific topic: \underline{C}OVID-19 \underline{V}accine, \underline{C}limate \underline{C}hange, and \underline{C}yber\underline{s}ecurity, denoted as \textit{CV}, \textit{CC} and \textit{CS}, respectively. These datasets were collected from two social media platforms to ensure diversity in the content and structural characteristics of the posts.

\vspace{-2mm}
\paragraph{COVID-19 Vaccine (\textit{CV}).}
We collected tweets related to COVID-19 vaccines using Wildfire~\cite{wildfire-wsdm24demo}.
For this dataset, we targeted tweets containing various keyword variations related COVID-19 vaccines, such as ``covid19 vaccination,'' ``covid-19 vaccine,'' and ``covid vax.''
The data collection period spanned from January 1, 2023 to April 25, 2023, resulting in a total of 384,676 tweets. After applying claim detection described in Section~\ref{sec:claim_detection}, 232,368 tweets were retained for distinct claim identification. 

\vspace{-2mm}
\paragraph{Climate Change (\textit{CC)}.}
For the climate change dataset, we utilized CrowdTangle~\cite{crowdtangle}---a now-discontinued tool---to collect Facebook posts related to climate change. We retrieved posts containing the keyword ``climate change'' between January 1, 2024, and May 7, 2024, yielding a total of 229,913 posts. After applying claim detection, we retained 89,412 posts.

\vspace{-2mm}
\paragraph{Cybersecurity (\textit{CS}).}
We collected Facebook posts related to cybersecurity using CrowdTangle, with the keyword ``cybersecurity.'' The collection period also spans from January 1, 2024 to May 7, 2024. Initially, 107,905 posts were gathered, and after claim detection, 38,530 posts were retained.

\subsection{Implementation Details}

For the HDBSCAN clustering model, we set the minimum cluster size to 3 for the \textit{CV} dataset. The minimum cluster size for \textit{CC} and \textit{CS} datasets was set to 2 because they have fewer posts, and we hoped to avoid the majority of posts being classified as outliers. 
We also set a maximum cluster size of 3,000 to prevent the formation of overly large clusters.
We noticed some clusters share identical posts, the reason for which is that Sentence-BERT may generate slightly different vector embeddings for the same sentence~\cite{reimers2019sentence}. To avoid duplication, we only keep the same post once in the distinct claims $\mathcal{C}$. The final numbers of distinct claims for \textit{CV}, \textit{CC}, and \textit{CS} datasets are 8,103, 14,408, and 5,731, respectively. 

For taxonomy construction, we employed three LLMs, Zephyr~\cite{tunstall2023zephyr}, GPT-4o mini~\cite{openai2024gpt4omini}, and Gemini 2.0 Flash~\cite{geminiflash2025} for performance comparison. Zephyr is selected for its competitive performance in language understanding tasks among all $7$-billion-parameter LLMs~\cite{chiang2024chatbot}, while GPT-4o mini and Gemini 2.0 Flash are chosen for their balance of cost-efficiency and performance.
For each dataset, we randomly selected 100 distinct factual claims and annotated them with broad, medium, and detailed topics. From annotated factual claims, we then chose representative samples based on their frequency of occurrence to serve as learning examples for the LLMs.
These annotated claims were used to guide the LLMs in generating topics for the distinct claims $\mathcal{C}$ identified through clustering. 
All experiments were conducted on three A100 GPUs.

\subsection{Results}
\subsubsection{Clustering}

The statistics of the clusters are shown in Table~\ref{tab:cluster_statistics}.
We used Silhouette Coefficient~\cite{rousseeuw1987silhouettes} to evaluate the clusters' quality.  Since HDBSCAN was employed, the outlier cluster was excluded during the evaluation. The \textit{CV} dataset achieved the highest Silhouette score of 0.940, reflecting highly cohesive and well-separated clusters. Although the \textit{CC} and \textit{CS} datasets have lower scores of 0.488 and 0.554, these still indicate reasonably good cluster quality. 
The variation in Silhouette scores across different datasets can be attributed to differences in their characteristics, particularly the higher sparsity and greater length of Facebook posts, which make clustering more challenging.
These results suggest that the clustering method performs reasonably well across diverse datasets, even with different clustering configurations.

\begin{table}[h!]
\centering
\small
    \begin{tabular}{cccc}
    \toprule
    \textbf{Datasets} & \textbf{Posts} & \textbf{Clusters} & \textbf{Outliers} \\
    \midrule
    \textit{CV} & 232,368 & 10,995 & 25,962 \\
    \textit{CC} & 89,412 & 15,794 & 42,923 \\
    \textit{CS} & 38,530 & 7,398 & 15,946 \\
    \bottomrule
    \end{tabular}
\caption{Cluster statistics for different datasets.}
\vspace{-2mm}
\label{tab:cluster_statistics}
\end{table}

\subsubsection{Multi-level Topic Generation}
The statistics of the generated topics across datasets are presented in Table~\ref{tab:existing_topic}---the rows labeled ``w/'' under the method column. It is evident that Gemini and GPT-4o mini outperformed Zephyr in constraining the number of broad and medium topics to a narrower range. Gemini also demonstrated superior ability in limiting the number of detailed topics. 
However, we observed that some generated topics were associated with only a few factual claims. To enhance the taxonomy's readability, we consolidated broad topics that appear fewer than 50 times into a new broad category labeled ``Other.'' For medium and detailed topics, we retained those with occurrences exceeding 4 for medium topics and 3 for detailed topics, respectively. 
Topics with fewer occurrences were grouped under ``Other'' topic within their respective parent topics.
After merging, the topic statistics of each dataset are shown in Table~\ref{tab:topic_counts}. This approach limits the taxonomy to a more manageable size.

\begin{table}[h!]
    \centering
    \resizebox{\columnwidth}{!}{
        \begin{tabular}{cccccc}
        \toprule
        \textbf{Dataset} & \textbf{Model} & \textbf{Method} & \makecell{\textbf{Broad} \\ \textbf{Topic}} & \makecell{\textbf{Medium} \\ \textbf{Topic}} & \makecell{\textbf{Detailed} \\ \textbf{Topic}} \\
        \midrule
        \multirow{6}{*}{\textit{CV}} 
        & \multirow{2}{*}{Zephyr} &  w/o & 839  & 1585 & 6553 \\
        &  &  w/  & 125 (85.1$\%\downarrow$)  & 899 (43.3$\%\downarrow$)  & 6060 (7.5$\%\downarrow$) \\
        \cmidrule(lr){2-6}
        & \multirow{2}{*}{GPT-4o mini} &  w/o & 1028 & 2301 & 6399 \\
        &  &  w/  & \textbf{12 (98.8$\%\downarrow$)}   & \textbf{41 (98.2$\%\downarrow$)}  & 2073 (67.6$\%\downarrow$) \\
        \cmidrule(lr){2-6}
        & \multirow{2}{*}{Gemini} &  w/o & 309  & 1062 & 5828 \\
        &  &  w/  & 12 (96.1$\%\downarrow$)  & 50 (95.3$\%\downarrow$)  & \textbf{49 (99.2$\%\downarrow$)} \\
        \midrule
        \multirow{6}{*}{\textit{CC}} 
        & \multirow{2}{*}{Zephyr} & w/o  & 1046  & 3831 & 12977  \\
        &  &  w/  & 124 (88.1$\%\downarrow$) & 1092 (71.5$\%\downarrow$) & 7414 (42.9$\%\downarrow$) \\
        \cmidrule(lr){2-6}
        & \multirow{2}{*}{GPT-4o mini} &  w/o & 1668 & 4998 & 12638 \\
        &  &  w/  & \textbf{8 (99.5$\%\downarrow$)}   & 274 (94.5$\%\downarrow$) & 8722 (31.0$\%\downarrow$) \\
        \cmidrule(lr){2-6}
        & \multirow{2}{*}{Gemini} &  w/o & 385  & 2306 & 12360 \\
        &  &  w/  & 7 (98.2$\%\downarrow$)  & \textbf{42 (98.2$\%\downarrow$)}  & \textbf{573 (95.4$\%\downarrow$)} \\
        \midrule
        \multirow{6}{*}{\textit{CS}} 
        & \multirow{2}{*}{Zephyr} &  w/o & 377  & 1684 & 5340 \\
        &  &  w/  & 126 (66.6$\%\downarrow$) & 656 (61.0$\%\downarrow$) & 3376 (36.8$\%\downarrow$) \\
        \cmidrule(lr){2-6}
        & \multirow{2}{*}{GPT-4o mini} &  w/o & 688  & 2184 & 5335 \\
        &  &  w/  & \textbf{12 (98.3$\%\downarrow$)}  & 111 (94.9$\%\downarrow$) & 4887 (8.4$\%\downarrow$) \\
        \cmidrule(lr){2-6}
        & \multirow{2}{*}{Gemini} &  w/o & 225  & 1572 & 5206 \\
        &  &  w/  & 9 (96.0$\%\downarrow$)  & \textbf{70 (95.5$\%\downarrow$)}  & \textbf{760 (85.4$\%\downarrow$)} \\
        \bottomrule
        \end{tabular}
    }
    \caption{Topic counts for different datasets with and without prompting seed taxonomy.}
    \label{tab:existing_topic}
\end{table}

\begin{table}[h!]
    \centering
    \small
    % \resizebox{\columnwidth}{!}{
        \begin{tabular}{ccccc}
        \toprule
        \textbf{Dataset} & \textbf{Model} & \makecell{\textbf{Broad} \\ \textbf{Topic}} & \makecell{\textbf{Medium} \\ \textbf{Topic}} & \makecell{\textbf{Detailed} \\ \textbf{Topic}} \\
        \midrule
        \makecell{\textit{CV}} & \makecell{Zephyr \\ GPT-4o mini \\ Gemini} & \makecell{11\\ 8 \\ 9 } & \makecell{ 66 \\ 18 \\ 27 } & \makecell{114 \\ 110 \\ 12 }  \\
        \midrule
        \makecell{\textit{CC}} & \makecell{Zephyr \\ GPT-4o mini \\ Gemini} & \makecell{8\\ 7  \\ 7} & \makecell{ 146 \\ 46 \\ 22 } & \makecell{163 \\ 229 \\ 30 }  \\
        \midrule
        \textit{CS} & \makecell{Zephyr \\ GPT-4o mini \\ Gemini} & \makecell{10\\ 9  \\ 9 } & \makecell{ 48 \\ 25 \\ 32 } & \makecell{32 \\ 61 \\ 26 }  \\
        \bottomrule
        \end{tabular}
    % }
    \caption{Topic counts after merging less frequent topics.}
    \vspace{-2mm}
    \label{tab:topic_counts}
\end{table}

\subsection{Ablation Study}
\label{sec:ablation}
To evaluate the effectiveness of the presence of the seed taxonomy in the prompt, we conducted experiments with removing the seed taxonomy from the prompt. Specifically, we only provide the LLMs with instructions, learning examples, and the target factual claims. The total topic counts with and without seed taxonomy are shown in Table~\ref{tab:existing_topic}. We can see that adding the seed taxonomy to the prompt effectively restricts the taxonomy size, reducing the number of broad topics by up to 99.5\%.
\section{Evaluation}
\label{sec:evaluation}
We assessed \LLMTaxo\ from two perspectives: 1)~the quality of the generated taxonomy, and 2) the relevance of the generated topics to the factual claims.
We developed a set of evaluation metrics to assess these two aspects. 
We engaged both human evaluators and GPT-4~\cite{achiam2023gpt}  for the evaluation. The effectiveness of using LLMs for model performance evaluation has been validated by previous studies~\cite{fu-etal-2024-gptscore, liu2023g}.
Identical instructions and metrics were provided to both human and GPT-4 evaluators. They were instructed to rate each criterion on a scale from 1 (strongly disagree) to 5 (strongly agree), where 5 indicates the best quality and 1 the worst.

To evaluate the quality of taxonomies, we presented taxonomies generated by Zephyr, GPT-4o mini, and Gemini across the three datasets to the evaluators, resulting in a total of 9 taxonomies.
For the evaluation of the claim-topic suitability, we randomly selected 102 factual claims along with their corresponding lowest-level topics (i.e., leaf nodes of the taxonomy) generated by the three models from each dataset, resulting a total of 306 claim-topic pairs for evaluation. 
We exclusively evaluated the leaf node topics with claims, as the broader topics had already been assessed in taxonomy quality evaluation.
Note that not every  claim has three-level topics, as mentioned in Section~\ref{sec:taxonomy_construction}. Therefore, we focused our evaluation on the leaf node topics.
To mitigate bias, we shuffled claim-topic pairs from different models before presenting them to the evaluators. Both human evaluators and GPT-4 reviewed these pairs using our predefined metrics. Human evaluators were required to provide rationales for the scores. The evaluation prompt for GPT-4 is detailed in Appendix~\ref{sec:eval_prompt}.

\subsection{Taxonomy Evaluation}
\label{sec:taxonomy_eval}
To design the taxonomy evaluation metrics, we adopted the Goal Question Metric (GQM) approach~\cite{caldiera1994goal} and consulted existing metrics from \citet{kaplan2022introducing}. Due to differences in our tasks, we retained only the \textit{orthogonality} and \textit{completeness} metrics from \citet{kaplan2022introducing}. We further refined these metrics to better align with our objectives and introduced additional metrics tailored to our evaluation needs. Each metric is defined with a clear goal, a guiding question, and specified evaluation criteria. An overview of the metrics is provided below, with comprehensive details available in Appendix~\ref{sec:eval_metrics}. 

\vspace{-2mm}
\paragraph{Clarity.} The goal of this metric is to ensure that each topic label communicates its content effectively to avoid confusion. To assess whether the topic labels are clear, precise, and unambiguous, we evaluate precision, unambiguity, consistency, and accessibility. 

\vspace{-2mm}
\paragraph{Hierarchical coherence.} The goal is to ensure that the taxonomy's structure facilitates easy navigation and understanding by clearly organizing information from the most general to the most specific. To assess whether the taxonomy follows a clear and meaningful hierarchical structure, we evaluate gradational specificity, parent-child coherence, and consistency. 

\vspace{-2mm}
\paragraph{Orthogonality.} The goal is to maintain clear boundaries between topics, ensuring that each one captures a unique aspect of the overall topic domain. To assess whether the topics are well-differentiated and free from redundancy, we evaluate their distinctiveness and degree of overlap. 

\vspace{-2mm}
\paragraph{Completeness.} The goal of this metric is to cover as many areas of the topic domain as possible to ensure the taxonomy is comprehensive. To assess whether the taxonomy captures a broad and representative set of topics across different facets of the topic domain, we evaluate its domain coverage, depth, as well as balance.

Three human evaluators and GPT-4 were provided with evaluation instructions and the background of the taxonomy construction. Each of them rated the taxonomies based on the evaluation criteria. 
For each metric, we calculated the average score of the evaluation criteria for each human evaluator. We then computed the mean of these averages across the three human annotators. The evaluation scores are presented in Table~\ref{tab:evaluation_scores}. 

The results affirm the overall efficacy of \LLMTaxo, with taxonomies generally receiving high ratings (above 3.0 across all metrics). Notably, GPT-4o mini consistently outperformed Zephyr and Gemini, suggesting its greater suitability for taxonomy construction. 
Human evaluators tended to give higher scores than GPT-4, particularly for taxonomies generated by Zephyr and Gemini, indicating possibly stricter criteria or different interpretations of taxonomy quality by GPT-4. It is also reasonable to assume that GPT-4 shares greater similarity with GPT-4o mini than with Zephyr or Gemini. 

The models generally scored well on clarity and completeness, indicating their effectiveness in producing clear, precise and comprehensive topic labels. GPT-4o mini also scored higher in hierarchical coherence with the highest score of 4.7, suggesting it does well in structuring information from general to specific in a meaningful way. 
Orthogonality has slightly lower scores compared to others, particularly for Zephyr and Gemini, which indicates some overlap or less distinct boundaries between topics. We observed Gemini was more strongly influenced by the seed taxonomy and tended to generate topics closely aligned with it. This is reflected in the limited number of unique topics produced by Gemini, as shown in Table~\ref{tab:topic_counts} and Table~\ref{tab:existing_topic}. Such alignment negatively affected both orthogonality and completeness. 
To assess the agreement level among human evaluators, we computed the inter-rater reliability using Gwet's AC2~\cite{gwet2008computing} with quadratic weights, which yielded a score of 0.80.

\begin{table}[!ht]
    \centering
    \resizebox{\columnwidth}{!}{
        \begin{tabular}{ccccccccc}
            \toprule
            \textbf{Dataset} & \textbf{Evaluator} & \textbf{Model} & \textbf{M\textsubscript{1}} & \textbf{M\textsubscript{2}} & \textbf{M\textsubscript{3}} & \textbf{M\textsubscript{4}} & \textbf{Ac} & \textbf{Gr} \\
            \midrule
            \multirow{6}{*}{CV} 
            & \multirow{3}{*}{Human} & Zephyr & 4.2 & 4.0 & 3.3 & 4.3 & 4.7 & 3.6 \\
            &  & GPT-4o mini & 4.3 & 4.2 & 3.8 & 4.7 & 4.4 & 4.6 \\
            &  & Gemini & 4.4 & 3.9 & 3.5 & 3.9 & 4.4 & 4.6 \\
            \cmidrule(lr){2-9}
            & \multirow{3}{*}{GPT-4} & Zephyr & 3.5 & 3.7 & 3.0 & 3.7 & 4.0 & 4.6 \\
            &  & GPT-4o mini & 4.3 & 4.3 & 4.0 & 4.7 & 4.0 & 4.4\\
            &  & Gemini & 3.4 & 3.0 & 3.0 & 3.3 & 4.0 & 4.4 \\
            \midrule
            \multirow{6}{*}{CC} 
            & \multirow{3}{*}{Human} & Zephyr & 4.6 & 4.2 & 3.9 & 4.6 & 3.3 & 3.9\\
            &  & GPT-4o mini & 4.7 & 4.7 & 4.2 & 4.8 & 4.4 & 4.5\\
            &  & Gemini & 4.3 & 3.9 & 3.3 & 4.7 & 4.4 & 4.6 \\
            \cmidrule(lr){2-9}
            & \multirow{3}{*}{GPT-4} & Zephyr & 3.5 & 3.3 & 4.0 & 4.0 & 3.4 & 3.4\\
            &  & GPT-4o mini & 4.3 & 4.0 & 4.5 & 4.3 & 3.8 & 4.5\\
            &  & Gemini & 3.8 & 3.7 & 3.5 & 3.7 & 3.8 & 4.4 \\
            \midrule
            \multirow{6}{*}{CS} 
            & \multirow{3}{*}{Human} & Zephyr & 4.3 & 3.9 & 3.7 & 4.4 & 3.8 & 4.0\\
            &  & GPT-4o mini & 4.5 & 4.0 & 4.0 & 4.6 & 4.7 & 4.6\\
            &  & Gemini & 4.6 & 3.6 & 3.5 & 3.9 & 4.8 & 4.6 \\
            \cmidrule(lr){2-9}
            & \multirow{3}{*}{GPT-4} & Zephyr & 3.3 & 3.0 & 3.0 & 3.3 & 3.7 & 4.0\\
            &  & GPT-4o mini & 4.3 & 4.7 & 3.5 & 4.7 & 4.0 & 4.9\\
            &  & Gemini & 3.0 & 3.0 & 3.5 & 3.3 & 4.0 & 4.9 \\
            \bottomrule
        \end{tabular}
    }
    \caption{Taxonomy and claim-topic pairs evaluation scores for different models across datasets. In the table, $M_{1}$, $M_{2}$, $M_{3}$, $M_{4}$, Ac, and Gr represent Clarity, Hierarchical Coherence, Orthogonality, Completeness, Accuracy, and Granularity, respectively.}
    \vspace{-2mm}
    \label{tab:evaluation_scores}
\end{table}

\vspace{-2mm}
\subsection{Claim-Topic Evaluation}

To assess how well the factual claims are aligned with the topics assigned to them from the LLM-generated taxonomy, we performed evaluation on two aspects: \textit{accuracy} and \textit{granularity}.
As Section~\ref{sec:taxonomy_eval} already evaluated the taxonomy itself, which includes the relationships among broad, medium, and detailed topics, we only evaluated claims and their leaf node topics.
The detailed evaluation metrics are: 
\vspace{-2mm}
\paragraph{Accuracy.} 
This criterion assesses how accurately the leaf node topics reflect the content and context of the corresponding factual claims. This involves determining if the topics are relevant and if they correctly represent the underlying information in the claims without misinterpretation or error.
\vspace{-2mm}
\paragraph{Granularity.}
This criterion evaluates the specificity of the leaf node topics. This involves determining whether the topics are detailed enough to uniquely categorize and differentiate between factual claims, yet broad enough to maintain practical applicability over multiple claims. 

Nine human evaluators participated in the claim-topic evaluation. They were divided into three groups, each consisting of three individuals. Each group assessed the same set of 306 claim-topic pairs, comprising 34 pairs per model for each dataset. 
After completing the evaluations, we calculated the average scores for each model across the datasets, as presented in Table~\ref{tab:evaluation_scores}. 

Overall, GPT-4o mini and Gemini demonstrated stronger performance in both accuracy and granularity compared to Zephyr, particularly on the \textit{CS} dataset, where they achieved a granularity score of 4.9. 
According to the feedback from human evaluators, Zephyr often produced lengthy descriptions for detailed topics, occasionally repeating the content of the social media posts, which contributed to its lower granularity scores. 
However, Zephyr showed slightly higher accuracy in the \textit{CV} dataset in the human evaluation, indicating it may perform better with shorter texts such as tweets. 
In contrast, Facebook posts (\textit{CC} and \textit{CS}) likely provided more verbose and detailed content, which may have benefited GPT-4o mini and Gemini.

To measure the inter-rater reliability, we computed Gwet's AC2 scores for each group. The average scores of the three groups were 0.67 for accuracy and 0.75 for granularity.

\subsection{Error Analysis}
Based on the feedback from the human evaluators, errors in the taxonomies can be categorized into three types: 1)~overlapping of topics, 2)~lack of specificity in topic labels, and 3)~generation of noisy data.
For criteria receiving lower scores in taxonomy evaluation, evaluators highlighted issues caused by overlapping and similar topic labels, such as ``\textit{Vaccine Mandates}'' versus ``\textit{COVID-19 Vaccine Mandates}'' and ``\textit{Cybersecurity Levy on Transactions}'' versus ``\textit{Cybersecurity Levy on Bank Transactions}.'' Additionally, there are ambiguous labels, such as ``\textit{Lawsuits}'' and ``\textit{Cybersecurity},'' which lack specificity. 
In the evaluation of claim-topic pairs, some detailed topics directly mirrored the factual claims. In addition, the LLMs may produce irrelevant outputs, such as detailed topics labeled ``\textit{not mentioned in the given post}.'' These issues highlight areas where LLMs' performance can be improved to enhance the accuracy and relevance of the taxonomy, as well as the necessity of post-processing the generated topics.

\section{Conclusion}
In this study, we introduced \LLMTaxo, a novel framework that leverages LLMs to construct taxonomies of factual claims from social media.Through evaluations across three distinct datasets, our approach demonstrated effectiveness in organizing claims into hierarchical structures with broad, medium, and detailed topics. The results highlight the framework’s potential in reducing redundancy, improving information accessibility, and assisting users such as researchers and fact-checkers in navigating online factual claims.

%---------End Main Paper---------

\section*{Limitations}

The framework's reliance on LLMs such as Zephyr and GPT-4o mini introduces potential limitations. These models, while powerful, may generate inconsistent or irrelevant topics for certain factual claims, which can impact the overall quality and reliability of the taxonomy. Despite efforts to automate the process, manual intervention remains necessary for refining and annotating learning examples. This introduces subjectivity, which may affect the consistency of the taxonomy.

Scalability poses an additional challenge. While \LLMTaxo\ is designed to be adaptable for different topic domains, its performance may be constrained when applied to larger datasets with a broader range of topics. The computational demands associated with LLM-based topic generation could also limit its feasibility for large-scale applications.

\section*{Ethics and Risks}
The deployment of \LLMTaxo\ could potentially raise ethical considerations and risks. One key concern is bias in topic generation. Since LLMs are trained on vast amounts of pre-existing data, they may inadvertently reflect biases present in those sources. This can lead to biased topic categorizations, affecting the neutrality and fairness of the taxonomy. Identifying and mitigating such biases is crucial to maintaining objectivity.

Privacy considerations must also be addressed, as social media posts used in the study may contain personal information. Steps have been taken to anonymize and aggregate the data,
but ongoing vigilance is required to ensure compliance with ethical guidelines and protect individual privacy.

The framework could also be susceptible to misuse. Malicious actors may attempt to manipulate the taxonomy to frame narratives that align with specific agendas. To counteract this risk, transparency in methodology and responsible use of the framework should be prioritized.

Finally, the taxonomy has the potential to influence public perception of factual claims on social media. Care must be taken to ensure that it presents an accurate, balanced, and comprehensive view of claims, avoiding any unintentional misrepresentation of content.

\section*{Acknowledgments}
This work is partially supported by the United States National Science Foundation Award 2346261. We acknowledge the Texas Advanced Computing Center (TACC) for providing HPC resources that have contributed to the results reported in this paper.
We are grateful to the human evaluators---Jacob Devasier, 
Abhishek Divakar Goudar, 
Juan Guajardo Gutierrez, 
Akshay Kumar Rayapet Madhusudhan,  
Alankrit Moses, 
Sai Sandeep Naraparaju, 
Jeremiah Pitts, 
Theodora Toutountzi, 
Mai Tran, 
Qing Wang,
Devin Wingfield---for their thorough and dedicated work in taxonomy quality evaluation. 

\bibliography{custom, chengkai-bib}

\begin{thebibliography}{55}
\expandafter\ifx\csname natexlab\endcsname\relax\def\natexlab#1{#1}\fi

\bibitem[{Achiam et~al.(2023)Achiam, Adler, Agarwal, Ahmad, Akkaya, Aleman, Almeida, Altenschmidt, Altman, Anadkat et~al.}]{achiam2023gpt}
Josh Achiam, Steven Adler, Sandhini Agarwal, Lama Ahmad, Ilge Akkaya, Florencia~Leoni Aleman, Diogo Almeida, Janko Altenschmidt, Sam Altman, Shyamal Anadkat, et~al. 2023.
\newblock Gpt-4 technical report.
\newblock \emph{arXiv preprint arXiv:2303.08774}.

\bibitem[{A{\"\i}meur et~al.(2023)A{\"\i}meur, Amri, and Brassard}]{aimeur2023fake}
Esma A{\"\i}meur, Sabrine Amri, and Gilles Brassard. 2023.
\newblock Fake news, disinformation and misinformation in social media: a review.
\newblock \emph{Social Network Analysis and Mining}, 13(1):30.

\bibitem[{Allcott et~al.(2019)Allcott, Gentzkow, and Yu}]{allcott2019trends}
Hunt Allcott, Matthew Gentzkow, and Chuan Yu. 2019.
\newblock Trends in the diffusion of misinformation on social media.
\newblock \emph{Research \& Politics}, 6(2):2053168019848554.

\bibitem[{Blei et~al.(2003)Blei, Ng, and Jordan}]{blei2003latent}
David~M Blei, Andrew~Y Ng, and Michael~I Jordan. 2003.
\newblock Latent dirichlet allocation.
\newblock \emph{Journal of machine Learning research}, 3(Jan):993--1022.

\bibitem[{Brown et~al.(2020)Brown, Mann, Ryder, Subbiah, Kaplan, Dhariwal, Neelakantan, Shyam, Sastry, Askell et~al.}]{brown2020language}
Tom Brown, Benjamin Mann, Nick Ryder, Melanie Subbiah, Jared~D Kaplan, Prafulla Dhariwal, Arvind Neelakantan, Pranav Shyam, Girish Sastry, Amanda Askell, et~al. 2020.
\newblock Language models are few-shot learners.
\newblock \emph{Advances in neural information processing systems}, 33:1877--1901.

\bibitem[{Caldiera and Rombach(1994)}]{caldiera1994goal}
Victor R Basili1~Gianluigi Caldiera and H~Dieter Rombach. 1994.
\newblock The goal question metric approach.
\newblock \emph{Encyclopedia of software engineering}, pages 528--532.

\bibitem[{Campello et~al.(2013)Campello, Moulavi, and Sander}]{campello2013density}
Ricardo~JGB Campello, Davoud Moulavi, and J{\"o}rg Sander. 2013.
\newblock Density-based clustering based on hierarchical density estimates.
\newblock In \emph{Pacific-Asia conference on knowledge discovery and data mining}, pages 160--172.

\bibitem[{Carrion et~al.(2019)Carrion, Onorati, D{\'\i}az, and Triga}]{carrion2019taxonomy}
Belen Carrion, Teresa Onorati, Paloma D{\'\i}az, and Vasiliki Triga. 2019.
\newblock A taxonomy generation tool for semantic visual analysis of large corpus of documents.
\newblock \emph{Multimedia Tools and Applications}, 78:32919--32937.

\bibitem[{Chen et~al.(2023)Chen, Yi, and Varr{\'o}}]{chen2023prompting}
Boqi Chen, Fandi Yi, and D{\'a}niel Varr{\'o}. 2023.
\newblock Prompting or fine-tuning? a comparative study of large language models for taxonomy construction.
\newblock In \emph{2023 ACM/IEEE International Conference on Model Driven Engineering Languages and Systems Companion (MODELS-C)}, pages 588--596.

\bibitem[{Chen et~al.(2020)Chen, Lin, and Klein}]{chen2020constructing}
Catherine Chen, Kevin Lin, and Dan Klein. 2020.
\newblock Constructing taxonomies from pretrained language models.
\newblock \emph{arXiv preprint arXiv:2010.12813}.

\bibitem[{Chiang et~al.(2024)Chiang, Zheng, Sheng, Angelopoulos, Li, Li, Zhang, Zhu, Jordan, Gonzalez, and Stoica}]{chiang2024chatbot}
Wei-Lin Chiang, Lianmin Zheng, Ying Sheng, Anastasios~Nikolas Angelopoulos, Tianle Li, Dacheng Li, Hao Zhang, Banghua Zhu, Michael Jordan, Joseph~E. Gonzalez, and Ion Stoica. 2024.
\newblock Chatbot arena: An open platform for evaluating llms by human preference.

\bibitem[{{CrowdTangle}(2024)}]{crowdtangle}
{CrowdTangle}. 2024.
\newblock \href {https://transparency.meta.com/researchtools/other-datasets/crowdtangle} {{CrowdTangle: A Public Insights Tool}}.
\newblock Accessed: 2024-06-15.

\bibitem[{DeepMind(2025)}]{geminiflash2025}
Google DeepMind. 2025.
\newblock \href {https://deepmind.google/models/gemini/flash-lite/} {Gemini 2.0 flash lite}.
\newblock Accessed: 2025-05-26.

\bibitem[{Del~Vicario et~al.(2016)Del~Vicario, Bessi, Zollo, Petroni, Scala, Caldarelli, Stanley, and Quattrociocchi}]{del2016spreading}
Michela Del~Vicario, Alessandro Bessi, Fabiana Zollo, Fabio Petroni, Antonio Scala, Guido Caldarelli, H~Eugene Stanley, and Walter Quattrociocchi. 2016.
\newblock The spreading of misinformation online.
\newblock \emph{Proceedings of the national academy of Sciences}, 113(3):554--559.

\bibitem[{Dumais(2004)}]{dumais2004latent}
Susan~T Dumais. 2004.
\newblock Latent semantic analysis.
\newblock \emph{Annual Review of Information Science and Technology (ARIST)}, 38:189--230.

\bibitem[{Durham et~al.(2023)Durham, Chowdhury, and Alzarrad}]{durham2023unveiling}
James Durham, Sudipta Chowdhury, and Ammar Alzarrad. 2023.
\newblock Unveiling key themes and establishing a hierarchical taxonomy of disaster-related tweets: A text mining approach for enhanced emergency management planning.
\newblock \emph{Information}, 14(7):385.

\bibitem[{Fu et~al.(2024)Fu, Ng, Jiang, and Liu}]{fu-etal-2024-gptscore}
Jinlan Fu, See-Kiong Ng, Zhengbao Jiang, and Pengfei Liu. 2024.
\newblock {GPTS}core: Evaluate as you desire.
\newblock In \emph{Proceedings of the 2024 Conference of the North American Chapter of the Association for Computational Linguistics: Human Language Technologies (Volume 1: Long Papers)}, pages 6556--6576.

\bibitem[{Geeraerts(2006)}]{geeraerts2006prototype}
Dirk Geeraerts. 2006.
\newblock Prototype theory.
\newblock \emph{Cognitive linguistics: Basic readings}, 34:141--165.

\bibitem[{Grootendorst(2022)}]{grootendorst2022bertopic}
Maarten Grootendorst. 2022.
\newblock Bertopic: Neural topic modeling with a class-based tf-idf procedure.
\newblock \emph{arXiv preprint arXiv:2203.05794}.

\bibitem[{Gwet(2008)}]{gwet2008computing}
Kilem~Li Gwet. 2008.
\newblock Computing inter-rater reliability and its variance in the presence of high agreement.
\newblock \emph{British Journal of Mathematical and Statistical Psychology}, 61(1):29--48.

\bibitem[{Hassan et~al.(2017)Hassan, Arslan, Li, and Tremayne}]{claimbuster-kdd17}
Naeemul Hassan, Fatma Arslan, Chengkai Li, and Mark Tremayne. 2017.
\newblock Toward automated fact-checking: Detecting check-worthy factual claims by {ClaimBuster}.
\newblock In \emph{Proceedings of the 23rd ACM SIGKDD International Conference on Knowledge Discovery and Data Mining (KDD)}, pages 1803--1812.

\bibitem[{Hook and Verdeja(2022)}]{hook2022social}
Kristina Hook and Ernesto Verdeja. 2022.
\newblock Social media misinformation and the prevention of political instability and mass atrocities.
\newblock Accessed: 2025-02-10.

\bibitem[{Huang et~al.(2020)Huang, Xie, Meng, Zhang, and Han}]{huang2020corel}
Jiaxin Huang, Yiqing Xie, Yu~Meng, Yunyi Zhang, and Jiawei Han. 2020.
\newblock Corel: Seed-guided topical taxonomy construction by concept learning and relation transferring.
\newblock In \emph{Proceedings of the 26th ACM SIGKDD International Conference on Knowledge Discovery \& Data Mining}, pages 1928--1936.

\bibitem[{Jung et~al.(2024)Jung, Lee, Woo, Baek, and Kim}]{jung2024expansive}
Hae~Sun Jung, Haein Lee, Young~Seok Woo, Seo~Yeon Baek, and Jang~Hyun Kim. 2024.
\newblock Expansive data, extensive model: Investigating discussion topics around llm through unsupervised machine learning in academic papers and news.
\newblock \emph{Plos one}, 19(5):e0304680.

\bibitem[{Kaplan et~al.(2022)Kaplan, K{\"u}hn, Hahner, Benkler, Keim, Fuch{\ss}, Corallo, and Heinrich}]{kaplan2022introducing}
Angelika Kaplan, Thomas K{\"u}hn, Sebastian Hahner, Niko Benkler, Jan Keim, Dominik Fuch{\ss}, Sophie Corallo, and Robert Heinrich. 2022.
\newblock Introducing an evaluation method for taxonomies.
\newblock In \emph{Proceedings of the 26th International Conference on Evaluation and Assessment in Software Engineering}, pages 311--316.

\bibitem[{Kumar and Shah(2018)}]{kumar2018false}
Srijan Kumar and Neil Shah. 2018.
\newblock False information on web and social media: A survey.
\newblock \emph{arXiv preprint arXiv:1804.08559}.

\bibitem[{Liu et~al.(2023)Liu, Iter, Xu, Wang, Xu, and Zhu}]{liu2023g}
Yang Liu, Dan Iter, Yichong Xu, Shuohang Wang, Ruochen Xu, and Chenguang Zhu. 2023.
\newblock G-eval: Nlg evaluation using gpt-4 with better human alignment.
\newblock \emph{arXiv preprint arXiv:2303.16634}.

\bibitem[{Miller(1995)}]{WordNet}
George~A. Miller. 1995.
\newblock Wordnet: A lexical database for english.
\newblock Princeton University.

\bibitem[{Moravec et~al.(2018)Moravec, Minas, and Dennis}]{moravec2018fake}
Patricia Moravec, Randall Minas, and Alan~R Dennis. 2018.
\newblock Fake news on social media: People believe what they want to believe when it makes no sense at all.
\newblock \emph{Kelley School of Business research paper}, (18-87).

\bibitem[{Mu et~al.(2024)Mu, Dong, Bontcheva, and Song}]{mu2024large}
Yida Mu, Chun Dong, Kalina Bontcheva, and Xingyi Song. 2024.
\newblock Large language models offer an alternative to the traditional approach of topic modelling.
\newblock \emph{arXiv preprint arXiv:2403.16248}.

\bibitem[{Muhammed~T and Mathew(2022)}]{muhammed2022disaster}
Sadiq Muhammed~T and Saji~K Mathew. 2022.
\newblock The disaster of misinformation: a review of research in social media.
\newblock \emph{International journal of data science and analytics}, 13(4):271--285.

\bibitem[{Najem and Hadi(2021)}]{najem2021semi}
Yasir~Abdalhamed Najem and Asaad~Sabah Hadi. 2021.
\newblock Semi-automatic ontology learning for twitter messages based on semantic feature extraction.
\newblock In \emph{New Trends in Information and Communications Technology Applications: 5th International Conference, NTICT 2021, Baghdad, Iraq, November 17--18, 2021, Proceedings 5}, pages 3--16.

\bibitem[{Navigli and Ponzetto(2010)}]{navigli2010babelnet}
Roberto Navigli and Simone~Paolo Ponzetto. 2010.
\newblock Babelnet: Building a very large multilingual semantic network.
\newblock In \emph{Proceedings of the 48th annual meeting of the association for computational linguistics}, pages 216--225.

\bibitem[{Ni et~al.(2024)Ni, Shi, Stammbach, Sachan, Ash, and Leippold}]{ni2024afacta}
Jingwei Ni, Minjing Shi, Dominik Stammbach, Mrinmaya Sachan, Elliott Ash, and Markus Leippold. 2024.
\newblock Afacta: Assisting the annotation of factual claim detection with reliable llm annotators.
\newblock \emph{arXiv preprint arXiv:2402.11073}.

\bibitem[{Ognyanova et~al.(2020)Ognyanova, Lazer, Robertson, and Wilson}]{ognyanova2020misinformation}
Katherine Ognyanova, David Lazer, Ronald~E Robertson, and Christo Wilson. 2020.
\newblock Misinformation in action: Fake news exposure is linked to lower trust in media, higher trust in government when your side is in power.
\newblock \emph{Harvard Kennedy School Misinformation Review}.

\bibitem[{OpenAI(2024)}]{openai2024gpt4omini}
OpenAI. 2024.
\newblock \href {https://openai.com/index/gpt-4o-mini-advancing-cost-efficient-intelligence/} {Gpt-4o-mini: Optimized mini version of gpt-4}.
\newblock Accessed: 2025-02-10.

\bibitem[{Pennycook et~al.(2018)Pennycook, Cannon, and Rand}]{pennycook2018prior}
Gordon Pennycook, Tyrone~D Cannon, and David~G Rand. 2018.
\newblock Prior exposure increases perceived accuracy of fake news.
\newblock \emph{Journal of experimental psychology: general}, 147(12):1865.

\bibitem[{Reimers(2019)}]{reimers2019sentence}
N~Reimers. 2019.
\newblock Sentence-bert: Sentence embeddings using siamese bert-networks.
\newblock \emph{arXiv preprint arXiv:1908.10084}.

\bibitem[{Rousseeuw(1987)}]{rousseeuw1987silhouettes}
Peter~J Rousseeuw. 1987.
\newblock Silhouettes: a graphical aid to the interpretation and validation of cluster analysis.
\newblock \emph{Journal of computational and applied mathematics}, 20:53--65.

\bibitem[{Sarkar et~al.(2023)Sarkar, Feng, and Karmaker\_Santu}]{sarkar2023zero}
Souvika Sarkar, Dongji Feng, and Shubhra~Kanti Karmaker\_Santu. 2023.
\newblock Zero-shot multi-label topic inference with sentence encoders and llms.
\newblock In \emph{Proceedings of the 2023 Conference on Empirical Methods in Natural Language Processing}, pages 16218--16233.

\bibitem[{Shah et~al.(2023)Shah, White, Andersen, Buscher, Counts, Das, Montazer, Manivannan, Neville, Ni et~al.}]{shah2023using}
Chirag Shah, Ryen~W White, Reid Andersen, Georg Buscher, Scott Counts, Sarkar Snigdha~Sarathi Das, Ali Montazer, Sathish Manivannan, Jennifer Neville, Xiaochuan Ni, et~al. 2023.
\newblock Using large language models to generate, validate, and apply user intent taxonomies.
\newblock \emph{arXiv preprint arXiv:2309.13063}.

\bibitem[{Suarez-Lledo and Alvarez-Galvez(2021)}]{suarez2021prevalence}
Victor Suarez-Lledo and Javier Alvarez-Galvez. 2021.
\newblock Prevalence of health misinformation on social media: systematic review.
\newblock \emph{Journal of medical Internet research}, 23(1):e17187.

\bibitem[{Tambini(2017)}]{tambini2017fake}
Damian Tambini. 2017.
\newblock Fake news: public policy responses.

\bibitem[{Tunstall et~al.(2023)Tunstall, Beeching, Lambert, Rajani, Rasul, Belkada, Huang, von Werra, Fourrier, Habib, Sarrazin, Sanseviero, Rush, and Wolf}]{tunstall2023zephyr}
Lewis Tunstall, Edward Beeching, Nathan Lambert, Nazneen Rajani, Kashif Rasul, Younes Belkada, Shengyi Huang, Leandro von Werra, Clémentine Fourrier, Nathan Habib, Nathan Sarrazin, Omar Sanseviero, Alexander~M. Rush, and Thomas Wolf. 2023.
\newblock Zephyr: Direct distillation of lm alignment.

\bibitem[{Wan et~al.(2024)Wan, Safavi, Jauhar, Kim, Counts, Neville, Suri, Shah, White, Yang et~al.}]{wan2024tnt}
Mengting Wan, Tara Safavi, Sujay~Kumar Jauhar, Yujin Kim, Scott Counts, Jennifer Neville, Siddharth Suri, Chirag Shah, Ryen~W White, Longqi Yang, et~al. 2024.
\newblock Tnt-llm: Text mining at scale with large language models.
\newblock In \emph{Proceedings of the 30th ACM SIGKDD Conference on Knowledge Discovery and Data Mining}, pages 5836--5847.

\bibitem[{Wu et~al.(2019)Wu, Morstatter, Carley, and Liu}]{wu2019misinformation}
Liang Wu, Fred Morstatter, Kathleen~M Carley, and Huan Liu. 2019.
\newblock Misinformation in social media: definition, manipulation, and detection.
\newblock \emph{ACM SIGKDD explorations newsletter}, 21(2):80--90.

\bibitem[{Xiao et~al.(2021)Xiao, Borah, and Su}]{xiao2021dangers}
Xizhu Xiao, Porismita Borah, and Yan Su. 2021.
\newblock The dangers of blind trust: Examining the interplay among social media news use, misinformation identification, and news trust on conspiracy beliefs.
\newblock \emph{Public Understanding of Science}, 30(8):977--992.

\bibitem[{Yang(2012)}]{yang2012constructing}
Hui Yang. 2012.
\newblock Constructing task-specific taxonomies for document collection browsing.
\newblock In \emph{Proceedings of the 2012 Joint Conference on Empirical Methods in Natural Language Processing and Computational Natural Language Learning}, pages 1278--1289.

\bibitem[{Zhang et~al.(2018)Zhang, Tao, Chen, Shen, Jiang, Sadler, Vanni, and Han}]{zhang2018taxogen}
Chao Zhang, Fangbo Tao, Xiusi Chen, Jiaming Shen, Meng Jiang, Brian Sadler, Michelle Vanni, and Jiawei Han. 2018.
\newblock Taxogen: Unsupervised topic taxonomy construction by adaptive term embedding and clustering.
\newblock In \emph{Proceedings of the 24th ACM SIGKDD International Conference on Knowledge Discovery \& Data Mining}, pages 2701--2709.

\bibitem[{Zhang et~al.(2024{\natexlab{a}})Zhang, Zhu, Zhang, Devasier, and Li}]{zhang-climatenlp24}
Haiqi Zhang, Zhengyuan Zhu, Zeyu Zhang, Jacob Devasier, and Chengkai Li. 2024{\natexlab{a}}.
\newblock Granular analysis of social media users' truthfulness stances toward climate change factual claims.
\newblock In \emph{Proceedings of the 1st Workshop on Natural Language Processing Meets Climate Change (ClimateNLP 2024)}, pages 233--240.

\bibitem[{Zhang et~al.(2024{\natexlab{b}})Zhang, Zhu, Zhang, Patel, Caraballo, Hennecke, and Li}]{wildfire-wsdm24demo}
Zeyu Zhang, Zhengyuan Zhu, Haiqi Zhang, Foram Patel, Josue Caraballo, Patrick Hennecke, and Chengkai Li. 2024{\natexlab{b}}.
\newblock Wildfire: {A} twitter social sensing platform for layperson.
\newblock In \emph{{Proceedings of the 17th ACM International Conference on Web Search and Data Mining (WSDM), demonstration description}}, pages 1106--1109.

\bibitem[{Zhao and Tsang(2022)}]{zhao2022people}
Xinyan Zhao and Stephanie Tsang. 2022.
\newblock How people process different types of misinformation on social media: A taxonomy based on falsity level and evidence type.
\newblock \emph{Available at SSRN 4259593}.

\bibitem[{Zhou et~al.(2015)Zhou, Zhao, and Lu}]{zhou2015impact}
Cangqi Zhou, Qianchuan Zhao, and Wenbo Lu. 2015.
\newblock Impact of repeated exposures on information spreading in social networks.
\newblock \emph{PloS one}, 10(10):e0140556.

\bibitem[{Zhu et~al.(2022)Zhu, Zhang, Patel, and Li}]{ratsd-compjour22}
Zhengyuan Zhu, Zeyu Zhang, Foram Patel, and Chengkai Li. 2022.
\newblock Detecting stance of tweets toward truthfulness of factual claims.
\newblock In \emph{Proceedings of the 2022 Computation+Journalism Symposium}.

\bibitem[{Zhu et~al.(2025)Zhu, Zhang, Zhang, and Li}]{ratsd-naacl25}
Zhengyuan Zhu, Zeyu Zhang, Haiqi Zhang, and Chengkai Li. 2025.
\newblock {RATSD}: Retrieval augmented truthfulness stance detection from social media posts toward factual claims.
\newblock In \emph{Findings of the Association for Computational Linguistics: NAACL 2025}, page 3366–3381.

\end{thebibliography}

\appendix

% \section{Appendix}

\begin{figure*}[ht]
    \centering    \includegraphics[width=0.8\textwidth]{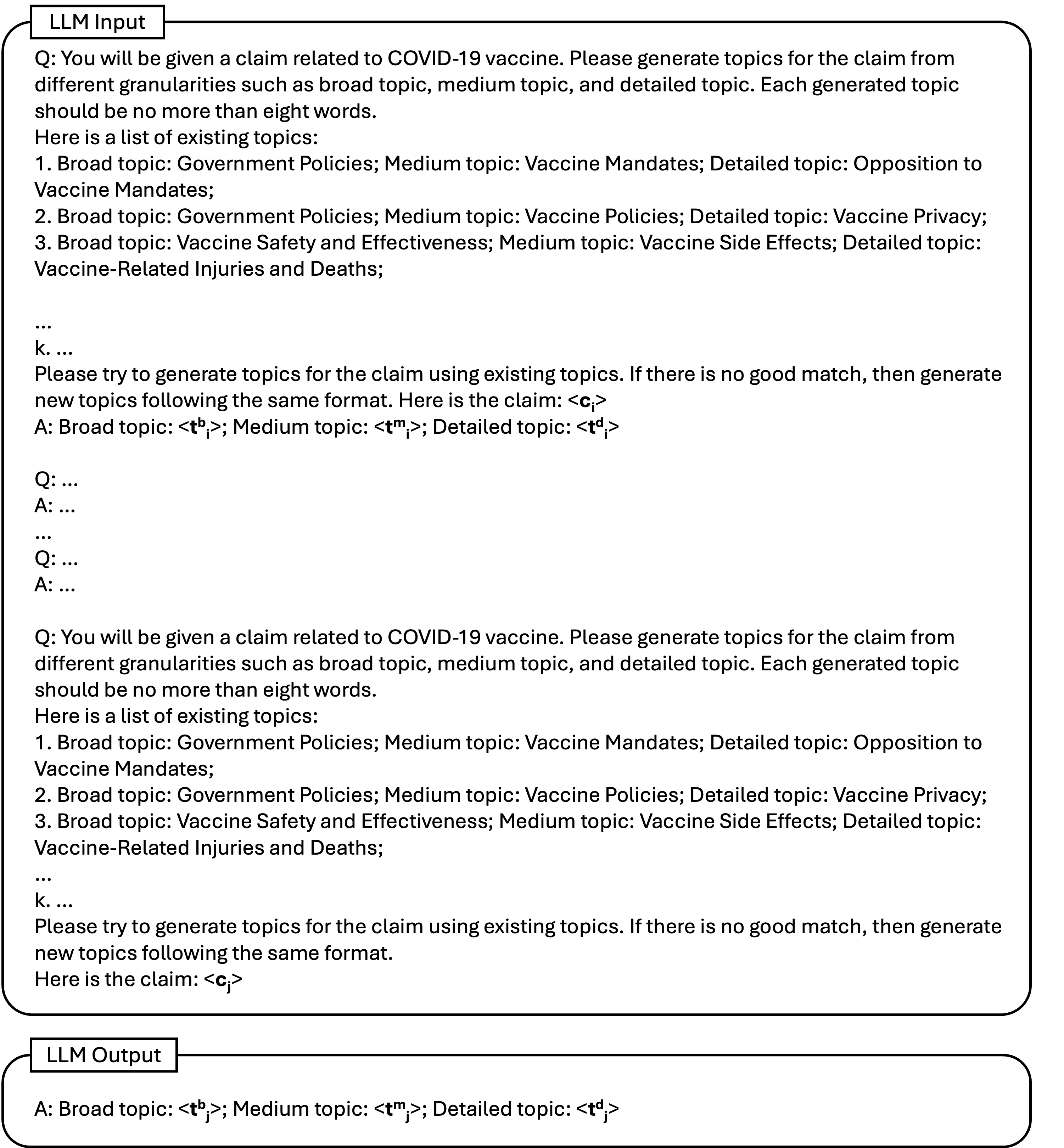}
    \caption{Prompt used to generate topics for each claim.}
    \label{fig:prompt}
\vspace{-3mm}
\end{figure*}

\section{Prompt for Topic Generation and Sample Results}
\label{sec:prompt}
\vspace{-1mm}
There are $k$ learning examples used to guide the LLMs in generating a broad topic, a medium topic, and a detailed topic for each factual claim, as shown in Figure~\ref{fig:prompt}. Each prompt example contains the list of topic tuples from the annotated factual claims in the $k$ learning examples (i.e., the seed taxonomy), a factual claim, a question asking the LLMs to generate broad, medium, and detailed topics for the claim, and the answer to the question. In the question, the LLMs are instructed to prioritize generating topics from the existing topics in the seed taxonomy. If none of the existing topics align well with the claim, the LLMs are then directed to generate new topics. 
This instruction ensures that the LLMs produce a limited number of topics. 
This prompt is iterated through all the factual claims to generate topics for them. 
A sample of the generated results from the three datasets (\textit{CV}, \textit{CC}, and  \textit{CS}) are shown in Table~\ref{tab:topics_results}.\vspace{-1mm}

\begin{table*}[ht]
\centering
\small
\begin{tabular}{p{1.6cm}p{6.3cm}p{1.7cm}p{1.5cm}p{2.3cm}}
\toprule
\textbf{Dataset} & \textbf{Claim} & \textbf{Broad Topic} & \textbf{Medium Topic} & \textbf{Detailed Topic} \\ \midrule
COVID-19 Vaccine & John Stockton boldly suggests `thousands' of pro athletes died after Covid vaccine shot https://t.co/nXbt6Apm2q via @marca & Vaccine Safety and Effectiveness & Vaccine Side Effects & Vaccine-Related Injuries and Deaths \\ \midrule
COVID-19 Vaccine & A lot of people in ‘stage 4 cancer’ after \#Covid \#Vaccine https://t.co/z0YAqGgQrL & Vaccine Safety and Effectiveness & Vaccine Side Effects & Cancer Side Effect \\ \midrule
Climate Change & Climate change is an existential threat to humanity. On Earth Day and every day, we remain committed to taking the most aggressive climate action ever. & Activism and Public Awareness & Climate Advocacy & Aggressive Climate Action \\ \midrule
Climate Change & Climate change causes Dry spell in Kashmir, The weather in Kashmir is warmer than Delhi and Chandigarh, No snow rain in Kashmir During Chillai Kalan & Environmental Impact & Global Warming & Climate Change Effects in Kashmir \\ \midrule
Cybersecurity & CBN Exempts 16 Items from Cybersecurity Levy...including Salary, Loans, Pension, Donations & Policies and Governance & Government Regulations & Cybersecurity Levy Exemptions \\ \midrule
Cybersecurity & Streaming giant Roku has recently been targeted by a pair of cyberattacks, and the company confirmed over a half million Roku accounts were compromised. & Threats & Cyberattacks & Roku Account Compromise \\ \bottomrule
\end{tabular}
\caption{Factual claims and their topics generated by GPT-4o mini in different datasets.}
\label{tab:topics_results}
\end{table*}

\vspace{-0.3mm}
\section{GPT-4 Prompt for Evaluation}
\label{sec:eval_prompt}
\vspace{-1mm}
\subsection{Prompt for Evaluating Taxonomy}
\vspace{-1mm}
I used LLMs to construct taxonomy and now I need to evaluate the taxonomy. I created some metrics to evaluate it. The [Taxonomy file name] uploaded contains the taxonomy with three-level topics. Please use the metrics in the [Metrics file name] to evaluate the taxonomy in [Taxonomy file name]. Please read each metric and understand them clearly, and then rate the metrics from 1-5, where 5 is the highest quality and 1 is the lowest. Please also provide justifications for your score and ignore the topic ``Other'' during evaluation.

\subsection{Prompt for Evaluating Claim-Topic Pairs}
I used LLMs to generate topics from three levels for factual claims. Now I need to evaluate ONLY the detailed topics from two aspects: **accuracy** and **granularity**. 
Here are the two aspects:
Accuracy: This criterion assesses how accurately the leaf node topics reflect the content and context of the corresponding factual claims. This involves determining if the topics are relevant and if they correctly represent the underlying information without misinterpretation or error.
Granularity: This criterion evaluates the specificity of the leaf node topics. This involves determining whether the topics are detailed enough to uniquely categorize and differentiate between factual claims, yet broad enough to maintain practical applicability across multiple claims.
If there is no detailed topic for a claim then evaluate the medium topic. If there is no medium topic existing, then evaluate broad topic.

Please read the evaluation metrics carefully and evaluate the claim-topic pairs and give one score for accuracy and one score for granularity for each claim-topic pair. The score ranges from 1-5, with 5 being the best and 1 being the worst. 

<<EXAMPLES>>

<<EXAMPLE 1>>

Factual claim: I worked for 18 months to end Biden’s unscientific and unethical military COVID vaccine mandate. Thanks to your phone calls and letters, we gained 92 sponsors on HR 3860. Repeal of the mandate just became a reality with the signing of the NDAA. Now let’s end the other mandates. 

broad topic: Government Policies; medium topic: Vaccine Mandates; detailed topic: Opposition to Vaccine Mandates. 

Accuracy: 5. Granularity: 5.

<<EXAMPLE 2>>

Factual claim: Myocarditis is up TEN times due to the Covid Vaccine... Nearly 30 \% of young people have measurable cardiac injuries post-vaccine.. The CDC is LYING about this… 

broad topic: Vaccine Safety and Effectiveness; medium topic: Vaccine Side Effects; detailed topic: Myocarditis Side Effect 

Accuracy: 5. Granularity: 5.

<<EXAMPLE 3>>

Factual claim: Graphen oxide resonates at 26ghz microwaves from a 5G cell towers that’s in the COVID vaccine! You can neutralise the EMF and 5G radiation from mobile devices and detox from heavy metals. 

broad topic: Political and Societal Implications; medium topic: Conspiracy Theories

Accuracy: 5. Granularity: 5.

<<EXAMPLE 4>>

Factual claim: Study published in Dec. 2020 proved COVID Vaccines could cause Strokes, Alzheimer’s, Parkinson’s, Multiple Sclerosis, and Autoimmune Disorder – Is there any wonder why the Five Eyes; Europe have suffered 2 Million Excess Deaths in the past 2 years?

broad topic: Vaccine Safety and Effectiveness; medium topic: Scientific and Medical Discussions; detailed topic: Discussions about Strokes, Alzheimer's, Parkinson's, Multiple Sclerosis, and Autoimmune Disorder. 

Accuracy: 4. Granularity: 2.

<<EXAMPLE 5>>

Factual claim: 'The doctor said that the probable cause of her heart attack was the vaccine, but he was too scared to put that on the report.' South African politician Jay Naidoo reacts to the South African court being asked to conduct a judicial review of the Covid vaccine.

broad topic: Political and Societal Implications; medium topic: Vaccine Injury; detailed topic: Court Review of Covid Vaccine.

Accuracy: 2. Granularity: 5.

<<END EXAMPLES>>

Now, please evaluate the topics for the following claim-topic pairs and only provide the scores for accuracy and granularity separated by a comma. For example. 3, 4.

Claim: \{claim\}

Broad Topic: \{broad\_topic\}

Medium Topic: \{medium\_topic\}

Detailed Topic: \{detailed\_topic\}

\section{Evaluation Metrics}
\label{sec:eval_metrics}
This section provides a detailed explanation of the taxonomy evaluation metrics. 

\vspace{-2mm}
\paragraph{Clarity.}
Assess whether the topic labels are clear, precise, and unambiguous.

\textit{Purpose}: Ensure that each topic label communicates its content effectively to avoid confusion.

\textit{Evaluation Criteria}: 
\vspace{-2mm}
\begin{itemize}
\setlength{\itemsep}{-1.5mm}
    \item Precision: Each topic label uses specific and well-defined terms.
    \item Unambiguity: Topic labels should have only one interpretation, preventing misunderstanding.
    \item Consistency: Use of terminology is consistent across all levels of the taxonomy.
    \item Accessibility: Language is straightforward, avoiding jargon where possible unless it is standard within the covered domain.
\end{itemize}

\vspace{-2mm}
\paragraph{Hierarchical Coherence.} 
Assess whether the taxonomy follows a clear and meaningful hierarchical structure. 

\textit{Purpose}: Ensure that the taxonomy's structure facilitates easy navigation and understanding by clearly organizing information from the most general to the most specific.

\textit{Evaluation Criteria}: 
\vspace{-2mm}
\begin{itemize}
\setlength{\itemsep}{-1.5mm}
    \item Gradational Specificity: There is a logical progression from broader to more specific categories.
    \item Parent-Child Coherence: Parent-child relationships are well-formed, ensuring that child nodes logically belong to their parent nodes.
    \item Consistency: The hierarchy maintains consistent levels of detail throughout the taxonomy, ensuring that no topics are too broad or too narrow relative to others at the same level.
\end{itemize}

\vspace{-2mm}
\paragraph{Orthogonality.} 
Assess whether the topics are well-differentiated without duplication. 

\textit{Purpose}: Maintain distinct boundaries between topics to ensure that each topic captures unique aspects of the domain.

\textit{Evaluation Criteria}: 
\vspace{-2mm}
\begin{itemize}
\setlength{\itemsep}{-1.5mm}
    \item Distinctiveness: Topics at each level progressively add meaningful distinctions rather than just rephrasing broader topics.
    \item Non-overlap: For each topic, there is minimal to no overlap in the scope or content with other topics. Note that the topics with different parent topics are always different. For example, the medium topic ``Vaccine Safety'' under broad topic ``Public Opinion'' is essentially ``Public Opinion about Vaccine Safety'' and distinctly different from ``Vaccine Safety'' under ``Government Policies.'' To minimize redundancy, we use succinct descriptions that are sufficient to convey the distinct meaning of each topic.
    
\end{itemize}

\vspace{-2mm}
\paragraph{Completeness.} 
Assess whether the taxonomy captures a broad and representative set of topics across different aspects of the domain.

\textit{Purpose}: Cover as many areas of the topic to ensure the taxonomy is comprehensive.

\textit{Evaluation Criteria}: 
\vspace{-2mm}
\begin{itemize}
\setlength{\itemsep}{-1.5mm}
    \item Domain Coverage: The taxonomy covers a variety of significant aspects of the domain it represents.
    \item Depth: The taxonomy provides sufficient depth in each branch to capture nuanced distinctions within topics. 
    \item Balance: The topics are evenly distributed across the taxonomy. This involves assessing whether some branches are disproportionately detailed while others are underdeveloped, which could lead to an imbalance that might skew the taxonomy’s effectiveness and navigability.
\end{itemize}

Note that the intrinsic evaluation criteria of the metrics cannot completely eliminate overlap due to the inherent characteristics of taxonomy.

\section{Use of AI Assistants}
Some of our code was developed using GitHub Copilot, and the writing was polished using ChatGPT and Grammarly.

\section{Human Evaluators}
The human evaluators involved in the human evaluation are lab members of the research team.

\end{document}